\documentclass[runningheads]{llncs}
\usepackage[T1]{fontenc}
\usepackage{graphicx}
\usepackage{graphicx}
\usepackage{xcolor}
\usepackage{comment}
\usepackage[utf8]{inputenc}
\usepackage{multirow}
\usepackage{arydshln}
\usepackage{subfigure}
\usepackage{float}
\usepackage{array}
\usepackage{tabularx}
\usepackage{graphicx}
\usepackage{url}
\usepackage{adjustbox}
\usepackage{xspace}
\usepackage{xcolor}
\usepackage{comment}
\usepackage{caption} 
\usepackage{adjustbox} %smn
\usepackage{booktabs}
\usepackage{hyperref}
\PassOptionsToPackage{bookmarks=false}{hyperref}
\usepackage{nicematrix}
\restylefloat{table}
\usepackage{latexsym}
\usepackage{amsmath, amssymb}
\usepackage{type1cm}        % activate if the above 3 fonts are
\usepackage{comment}
\usepackage{makeidx}         % allows index generation
\usepackage{multicol}        % used for the two-column index
\usepackage[bottom]{footmisc}% places footnotes at page bottom
\usepackage{newtxtext}       % 
\usepackage[varvw]{newtxmath} 
\usepackage{type1cm}        % activate if the above 3 fonts are
\usepackage{makeidx}         % allows index generation
\usepackage{graphicx}        % standard LaTeX graphics tool
\usepackage{multicol}        % used for the two-column index
\usepackage[bottom]{footmisc}% places footnotes at page bottom
\usepackage{newtxtext}       % 
\usepackage[varvw]{newtxmath}       % selects Times Roman as basic 
\usepackage{color,array}
\usepackage{comment}
\usepackage{subcaption}
\usepackage{soul,xcolor} % for \sethlcolor{}
\usepackage{mdframed} %for med frame

\definecolor{highlightgreen}{rgb}{0.85, 1.0, 0.85} % Light green: correct
\definecolor{highlightorange}{rgb}{1.0, 0.9, 0.6} % Light orange: partially relevant
\definecolor{highlightred}{rgb}{1.0, 0.8, 0.8}  % Light red: hallucinated or irrelevant
\definecolor{highlightblue}{rgb}{0.85, 0.9, 1.0}  % Light blue: speculative or exaggerated
\definecolor{highlightyellow}{rgb}{1.0, 0.95, 0.5} % Slightly Deeper Yellow

\newcommand{\hlr}[1]{\sethlcolor{highlightred}\hl{#1}}

\makeindex 

\begin{document}
\title{Comparative Analysis of Abstractive Summarization Models for Clinical Radiology Reports}
\author{
Anindita Bhattacharya \inst{1} \and
Tohida Rehman \inst{1}\thanks{corresponding author}  \and
Debarshi Kumar Sanyal\inst{2} \and
Samiran Chattopadhyay\inst{3,1}
}

\authorrunning{Bhattacharya and Rehman et al.}
%\authorrunning{Bhattacharya, Rehman, Sanyal, and Chattopadhyay} 
\titlerunning{Bhattacharya and Rehman et al.}% Part of RIGHT running header
% If there are more than two authors, 'et al.' is used.%

\institute{
Jadavpur University, Kolkata, India.\\
\email{aninbhtry.01@gmail.com};
\email{tohidarehman.it@jadavpuruniversity.in}
\and
Indian Association for the Cultivation of Science, Kolkata, India.\\
\email{debarshi.sanyal@iacs.res.in}
\and
Techno India University, Kolkata, India.\\
\email{samirancju@gmail.com}
}

%\end{comment}

\maketitle
\begin{abstract}\unskip
    The `findings' section of a radiology report is often detailed and lengthy, whereas the `impression' section is comparatively more compact and captures key diagnostic conclusions. This research explores the use of advanced abstractive summarization models to generate the concise `impression' from the `findings' section of a radiology report. We have used the publicly available MIMIC-CXR dataset. A comparative analysis is conducted on leading pre-trained and open-source large language models, including T5-base, BART-base, PEGASUS-x-base, ChatGPT-4, LLaMA-3-8B, and a custom Pointer Generator Network with a coverage mechanism. To ensure a thorough assessment, multiple evaluation metrics are employed, including ROUGE-1, ROUGE-2, ROUGE-L, METEOR, and BERTScore. By analyzing the performance of these models, this study identifies their respective strengths and limitations in the summarization of medical text. The findings of this paper provide helpful information for medical professionals who need automated summarization solutions in the healthcare sector.


\end{abstract}
\section{Introduction}
The increasing volume of textual data, especially in specialized fields like medicine, presents a major challenge in extracting relevant information quickly and accurately. Radiology reports, for example, often contain complex and detailed observations that must be reviewed under time constraints. Abstractive summarization of these \textit{findings} into concise \textit{impressions} can assist medical professionals in delivering timely patient care. Table~\ref{tab:finding_impression_sample} shows an example of a \textit{Findings} and the corresponding \textit{Impression} in a radiology report. It helps doctors and radiologists quickly identify the core message of detailed reports, simplifies documentation, and supports timely decision-making. Additionally, it bridges communication gaps with patients and non-specialist staff, and facilitates research and training by making large volumes of radiology data easier to analyze.

\begin{table*}[h!]
\centering
\caption{Example of an input \textit{Findings} section and its corresponding ground-truth \textit{Impression} from the MIMIC-CXR dataset.}
\label{tab:finding_impression_sample}
\begin{adjustbox}{width=1.0\linewidth}
{\begin{tabular}{ |p{12.5 cm}|} \hline
\vspace{0.5 mm}
\textbf{Input \textit{Findings}}: \colorbox{blue!10}{``Single AP portable chest radiograph is obtained. Tracheostomy tube is present.} \colorbox{blue!10}{There is no pneumothorax or pleural effusion. There is a hazy veil-like opacity in the right upper} \colorbox{blue!10}{lung zone which may be consolidation, atelectasis or artifact. Heart size appears enlarged;} \colorbox{blue!10}{however, this may be technical due to AP view. Bony structures are intact.''} \\ \hline
\vspace{0.5 mm}
\textbf{Ground-Truth  \textit{Impression}}: \colorbox{green!10}{``Limited study with hazy opacity in the right upper and mid lungs} \colorbox{green!10}{which may be infectious in etiology, atelectasis or artifact.''} \\\hline
\end{tabular}}
\end{adjustbox}
\end{table*}	

Although Natural Language Processing (NLP) has advanced recently, particularly with the introduction of transformer-based models like T5 \cite{JMLR:v21:20-074}, BART \cite{lewis-etal-2020-bart}, and PEGASUS \cite{10.5555/3524938.3525989}, most summarization research has focused on general-purpose datasets like Wikipedia content or news items (CNN/DailyMail). These models, though powerful, are often not evaluated in domains like medicine, where accuracy, completeness, and domain-specific terminology are critical. Furthermore, while some studies have applied deep learning to medical summarization tasks, there is limited comparative analysis of different abstractive models specifically tailored to the radiology domain. The challenge lies in the domain adaptation of general models, handling medical terminology, ensuring factual consistency, and mitigating hallucinations in generated summaries.

The main goal of this study is to use a radiology dataset to perform a detailed comparative analysis of the most advanced abstractive summarization models, with an emphasis on fine-tuned and zero-shot variations. Models like T5, BART, PEGASUS, ChatGPT-4, LLaMA-3-8B, and Pointer Generator Networks will be evaluated in order to determine their advantages and disadvantages in generating clinically significant summaries. This paper aims to offer important insights into the performance and behavior of these models in a field that requires sensitivity and precision by methodically evaluating them within the framework of radiology reports. By shedding light on the usefulness of using sophisticated summarization models in healthcare applications, this study adds to the expanding corpus of research on clinical natural language processing.

The main contributions of this study are as follows:
\begin{enumerate}
 \item This study presents a comprehensive comparison of several prominent pre-trained and  large language models—namely, T5-base \cite{JMLR:v21:20-074}, BART-base~\cite{lewis-etal-2020-bart}, PEGASUS-x-base \cite{10.5555/3524938.3525989}, ChatGPT-4 \cite{achiam2023gpt}, LLaMA-3-8B \cite{touvron2023llama}, and a custom-designed Pointer Generator Network (PGN) with a coverage mechanism~\cite{see2017getTT} for the task of medical radiology summarization. The evaluation specifically examines their performance in summarizing \textit{findings} from the MIMIC-CXR~\cite{johnson2019mimic} dataset.
 
 \item The assessment leverages several well-established automated evaluation metrics, including variants of ROUGE (ROUGE-1, ROUGE-2, ROUGE-L) \cite{lin-2004-rouge}, METEOR \cite{banerjee-lavie-2005-meteor}, and BERTScore \cite{zhang2019bertscore}, to measure the informativeness and linguistic quality of the generated summaries.
 
 \item By analyzing the results across different models, this paper highlights their relative capabilities and challenges when applied to domain-specific summarization in radiology. It offers actionable insights into how each model performs when tasked with generating clinically meaningful summaries from complex, structured input texts.
\end{enumerate}

\section{Literature Review} 
Researchers have been exploring ways to automatically summarize text for a long time, and many approaches have been developed to tackle this challenge in Natural Language Processing. Ramesh Nallapati et al.~\cite{nallapati2016abstractive} introduced an abstractive summarization approach based on a sequence-to-sequence encoder-decoder model using RNNs. They proposed several improvements to overcome the limitations of the basic architecture, such as better handling of important words, capturing the sentence-to-word structure, and enabling the generation of rare or unseen words during training. Li et al.~\cite{li2017salience} introduced a summarization method that uses a recurrent generative decoder to improve output quality by capturing hidden structural patterns within the data. Duan et al.~\cite{duan2019contrastive} developed an abstractive summarization technique that applies contrastive attention to distinguish between essential and less important content, helping the model focus more effectively on relevant information. Rehman et al.~\cite{rehman2021automatic,rehman-etal-2022-named,rehman2023research,10172215,rehman2022analysis,rehman2022abstractive,rehman2024analysis} have extensively studied the performance of pre-trained models in summarization tasks across both open-domain content and scientific research papers.
Their research demonstrates how well-trained models handle massive amounts of textual input and generate concise summaries that may be used to improve search results. 

In recent years, radiology report summarization has drawn growing attention within the research community.  Hu et al.~\cite{hu2022graph} introduced a hybrid approach combining graph neural networks and contrastive learning to improve impression-level summarization of radiology reports, demonstrating strong performance on datasets like MIMIC-CXR ~\cite{johnson2019mimic} and OpenI \cite{demner2016preparing}. To improve the reliability of clinical summarization, the SPeC method~\cite{chuang2024spec} applies soft prompt-based calibration, which helps large language models generate more consistent and medically sound outputs. An iterative summarization framework~\cite{Ma_2024} leveraging ChatGPT and guided by human feedback across multiple rounds was designed to refine radiology report summaries, leading to improved precision and coherence in the generated \textit{impressions}. Wu et al.~\cite{wu2024radbartsum} introduced RadBARTsum, a domain-adapted version of BART fine-tuned on radiological texts to improve the generation of concise and context-aware summaries. Luo et al.~\cite{luo2025chatgpt} proposed a ChatGPT-based contrastive learning (CCL) approach for radiology report summarization, addressing data and exposure bias through fine-tuning, quality-guided negative sampling, and ChatGPT-driven augmentation, with improved performance on OpenI \cite{demner2016preparing} and MIMIC-CXR  ~\cite{johnson2019mimic} datasets. Sultan et al.~\cite{sultan2025sumgpt} proposed SumGPT, a multimodal framework combining T5 and a Vision Transformer, which outperforms existing models by effectively integrating visual and textual features. Yang et al.~\cite{yang2025aligning} investigated the use of reinforcement learning from AI feedback (RLAIF) to better align large language models with radiologist preferences in chest CT summarization, resulting in more clinically relevant outputs and offering a scalable alternative to manual evaluation.

\section{Datasets}
In this study, we used the \textbf{MIMIC-CXR}~\cite{johnson2019mimic} dataset, a publicly available collection of radiology reports. From the training set of 114,689 records, we selected the first 3,000 instances and split them into training and validation sets in an 80–20 ratio. For testing, 500 samples were taken from the original test split of 1,603 reports. 

\section{Methodology}

\subsection{Rationale for Model Selection}

\begin{figure*} [!htbp] 
\centering
\includegraphics[width=0.9\linewidth,height=7cm]{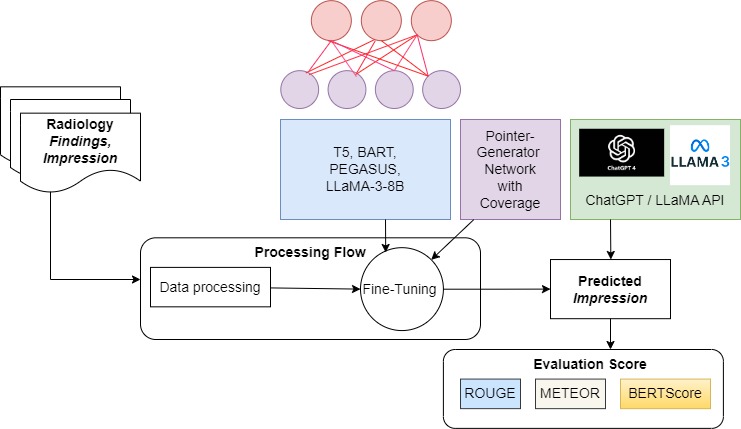}
\caption{Framework for radiology findings summarization using multiple model types.}
\label{fig:model_diagram}
\end{figure*}

To evaluate the effectiveness of different model types in clinical summarization, this study considers three key categories of language model architectures: encoder–decoder models (T5, BART), which generate structured summaries with strong contextual understanding; decoder-only models (GPT-4, LLaMA-3), known for producing fluent outputs through large-scale autoregressive training; and a custom Pointer Generator Network (PGN), which improves factual accuracy and interpretability by combining generation with a copying mechanism and coverage attention. Figure~\ref{fig:model_diagram} outlines the overall framework, including model-specific summarization pipelines, data flow, and evaluation metrics used to compare performance across all systems.

\subsection{Models Used}

\begin{enumerate}
    \item \textbf{T5-base:}
    T5 is a pre-trained transformer model that frames all NLP tasks as text-to-text problems, where both inputs and outputs are text sequences. This unified approach allows T5 model to handle a wide range of tasks, including summarization, by simply appending a task-specific prefix (e.g., ``summarize:") to the input text. T5's pre-training involves a span corruption objective, where spans of text are masked and the model learns to predict them. Its flexibility and ability to fine-tune effectively on task-specific datasets make it a strong contender for abstractive summarization tasks. T5-base model contains 220 million parameters.
    \item \textbf{BART-base:}
    BART combines the strengths of bidirectional encoding, which allows it to understand the full context of an input, with autoregressive decoding for effective text generation. Pre-trained as a denoising autoencoder, BART is adept at reconstructing corrupted input sequences, making it highly capable in text summarization tasks. Its encoder-decoder architecture ensures a deep understanding of input texts, resulting in coherent and concise summaries. BART-base contains 139 million parameters.
    \item \textbf{PEGASUS-X-base:}
    PEGASUS is explicitly designed for abstractive summarization and leverages a novel pre-training objective called Gap Sentence Generation (GSG). In GSG, key sentences from the input text are masked, and the model learns to predict them. This approach enhances PEGASUS's ability to focus on the most salient parts of a document. Its pre-training on large-scale datasets further boosts its performance on domain-specific summarization tasks. Pegasus-x-base contains around 223 million parameters.
    \item \textbf{ChatGPT-4:}
    ChatGPT, built on the GPT (Generative Pre-trained Transformer) architecture, demonstrates advanced conversational and text generation capabilities. Fine-tuned on diverse datasets, ChatGPT excels in generating coherent, contextually appropriate, and fluent summaries. Its capacity to adapt to various linguistic nuances and provide human-like text makes it particularly effective for abstractive summarization across different domains. Its precise architecture and training information are not publicly available.
    \item \textbf{LLaMA-3-8B:}
    LLaMA is a transformer-based language model designed to optimize performance on tasks requiring both language understanding and generation. We utilized the pre-trained {LLaMA-3-8B} model\footnote{\url{https://ai.meta.com/blog/meta-llama-3/}}, which comprises 8 billion parameters. The LLaMA model series~\cite{touvron2023llama} consists of decoder-only transformer-based large language models trained solely on publicly available datasets, distinguishing them from the GPT series.
    \item \textbf{Pointer Generator Network with Coverage Mechanism (PGNwithCOV):}
    The Pointer Generator Network combines the advantages of extractive and abstractive summarization. It incorporates a pointing mechanism that allows the model to copy words directly from the input text while simultaneously generating novel content. This feature ensures factual accuracy and preserves domain-specific terminology, which is critical for medical summaries. The addition of a coverage mechanism addresses the issue of repetition by maintaining a coverage vector that tracks attention over the input text. This tracking helps the model generate coherent and comprehensive summaries, making it highly suitable for radiology-specific summarization tasks.
\end{enumerate}

\section{Experimental Setup}

\subsection{Data Processing}
In the training set, the input texts from the \textit{findings} section ranged from 11 to 281 tokens, with an average length of approximately 57 tokens per instance. The corresponding target summaries from the \textit{impression} section ranged between 3 and 113 tokens, averaging around 16 tokens. For the test set, input lengths varied from 13 to 165 tokens (average 71 tokens), while the outputs ranged from 4 to 80 tokens, with an average of 20 tokens.

\subsection{Implementation Details}

As already mentioned, the encoder–decoder architecture of the \textbf{BART-base} model \footnote{\url{https://huggingface.co/facebook/bart-base}} was adapted for the summarization task. We also used the \textbf{Pegasus-x-base} model \footnote{\url{https://huggingface.co/google/pegasus-x-base}}, which is pre-trained with a gap-sentence generation objective and designed specifically for abstractive summarization. In addition, the \textbf{T5-base} model\footnote{\url{https://huggingface.co/t5-base}}, which treats all tasks as text-to-text problems, was applied due to its flexibility for generative tasks. For prompt-based evaluation, we used \textbf{ChatGPT-4} \footnote{\url{https://chatgpt.com/}} in a zero-shot setting. The \textbf{LLaMA-3-8B} model\footnote{\url{https://huggingface.co/meta-llama/Meta-Llama-3-8B}} was tested both in its base (zero-shot) and fine-tuned forms. Finally, a custom-built \textbf{Pointer Generator Network (PGN)} enhanced with a \textbf{Coverage Mechanism} was implemented for better factual accuracy and coverage control. All corresponding code is available on GitHub.\footnote{\url{https://github.com/tohidarehman/Summ-Radiology-Report}}

\paragraph{Prompt Template for ChatGPT-4 and LLaMA-3-8B LLMs:} The following prompt was used for ChatGPT-4 and LLaMA-3-8B to guide the summarization process:
%\begin{quote}
\begin{mdframed}[backgroundcolor=gray!10,linecolor=gray!50]
You are a medical AI assistant. Your role is to analyze the clinical <<findings>> provided below and produce a single, concise sentence for the <<impression>>. Do not include speculative or ambiguous information. Focus only on the key diagnostic and observational elements.
\end{mdframed}

For ChatGPT-4 in the prompt-based setting, we set the temperature to 0.5 to balance creativity and consistency. Top-$p$ was configured at 0.9, with frequency and presence penalties set to 0.2 and 0.1, respectively, to reduce repetition and promote lexical diversity. BART-base, T5-base, and PEGASUS-x-base were fine-tuned using a batch size of 8, a learning rate of {\tt 3e-5}, and trained for 5 epochs. LLaMA-3-8B was fine-tuned using a batch size of 2, a learning rate of {\tt 2e-4}, and also trained for 5 epochs. To accelerate training, we applied Low-Rank Adaptation (LoRA) with a rank of {\tt 16}, alpha {\tt 32}, and a dropout of  0.05. For all transformer models, the input and output sequence lengths were set to 512 and 150 tokens, respectively. The Pointer Generator Network (PGN) with coverage mechanism was also trained for 5 epochs with a fixed learning rate of {\tt 3e-5}, using the same sequence length settings. All experiments were conducted on \texttt{Google Colab}, with PGN training performed on a \textbf{v2-8 TPU} and the remaining models trained on \textbf{NVIDIA Tesla T4 GPUs}.

\begin{figure*}[!htbp]
\centering
\includegraphics[width=\linewidth, height=6cm]{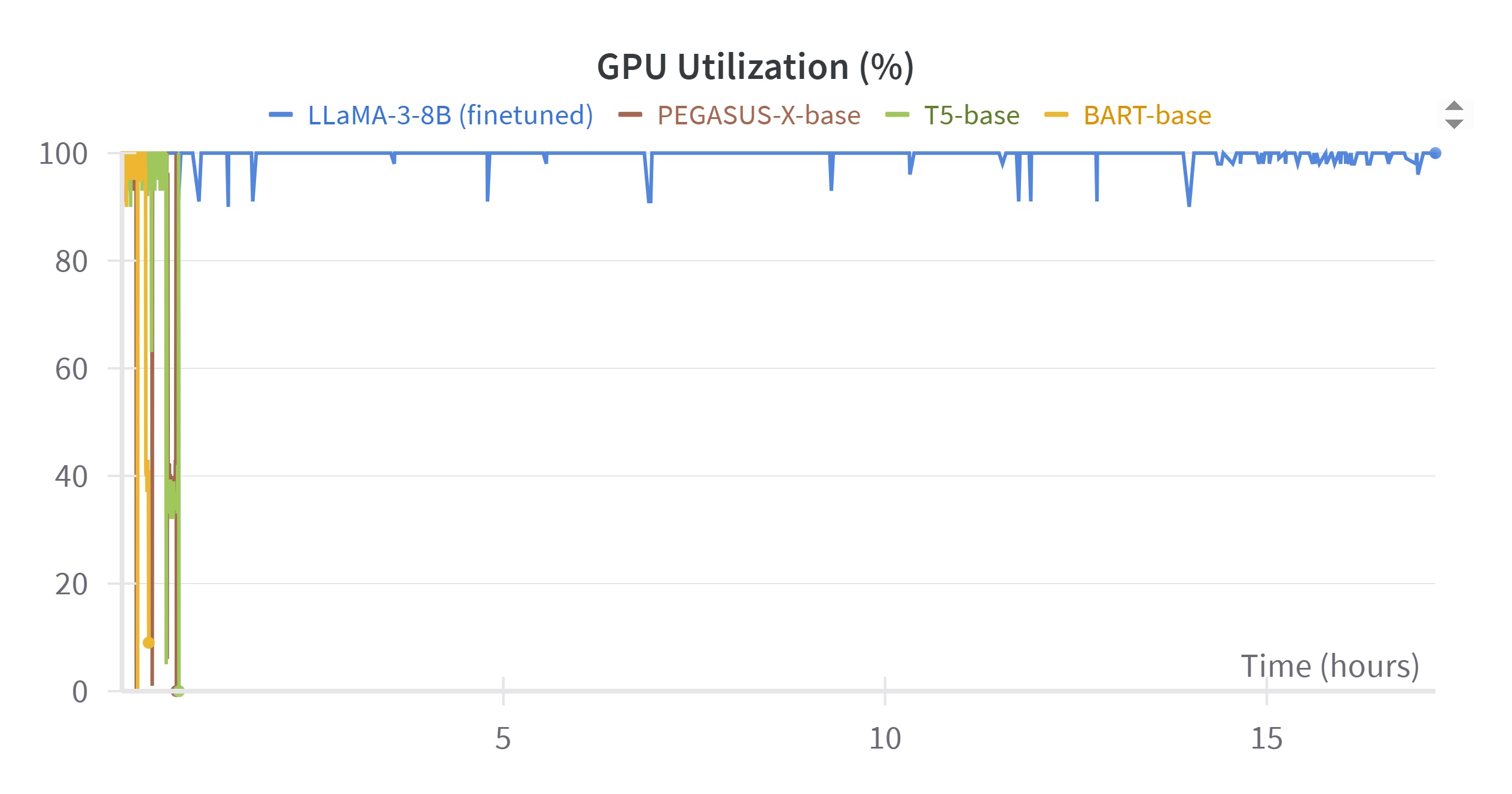}
\caption{GPU utilization of the pretrained language models (PLMs) and large language models (LLMs) used for radiology findings summarization.}
\label{fig:gpu-utilization}
\end{figure*}

During the fine-tuning of all models, we monitored memory and compute resource usage using the WandB tool \footnote{\url{https://wandb.ai/site}}. Among them, LLaMA-3 required the most GPU resources, showing longer training times and occasional spikes in memory access, whereas the other models maintained stable and efficient usage. 
Figure~\ref{fig:gpu-utilization} presents the GPU utilization across the pretrained and large language models used in this study.

\subsection{Evaluation Metrics}

The quality of the generated \textit{impression} is evaluated using ROUGE, METEOR, and BERTScore metrics. \textbf{ROUGE} \cite{lin-2004-rouge} assesses similarity between the model's output and the reference impression by comparing overlapping $n$-grams (ROUGE-1, ROUGE-2) and longest common subsequences (ROUGE-L), which help measure both lexical and structural alignment. 
\noindent \textbf{METEOR} \cite{banerjee-lavie-2005-meteor} evaluates matches based on unigrams, considering factors like stemming, synonyms, and word order, and applies a penalty for fragmented matches to reflect fluency. These metrics together offer a well-rounded view of how closely the generated impression mirrors the original clinical interpretation.

\noindent \textbf{BERTScore} \cite{zhang2019bertscore} evaluates how semantically close the generated impression is to the reference by using contextual embeddings from the BERT model. Unlike traditional $n$-gram-based metrics, it captures deeper meaning by comparing tokens in context, making it especially useful for evaluating abstractive outputs. Together with ROUGE and METEOR, BERTScore helps provide a comprehensive view of each model’s ability to produce accurate and meaningful clinical impressions.

\section{Results}

\subsection{Model Comparison}

We compared all models using ROUGE-1, ROUGE-2, ROUGE-L, BERTScore, and METEOR on the \textbf{MIMIC-CXR} dataset, with the results presented in Table~\ref{tab:results}. The fine-tuned \textbf{BART-base} model performed best overall (except for METEOR score), outperforming all other models in ROUGE-2 and ROUGE-L. For ROUGE-1 and BERTScore, both the fine-tuned \textbf{BART-base} and \textbf{T5-base} models attained equal performance. The fine-tuned \textbf{T5-base} model obtained the highest METEOR score. The fine-tuned \textbf{PEGASUS-x-base} demonstrated average performance, falling behind BART and T5 in several metrics. Despite being used in a zero-shot setting, \textbf{ChatGPT-4} produced competitive results, particularly for METEOR and BERTScore. Domain-specific fine-tuning led to clear improvements in the fine-tuned \textbf{LLaMA-3-8B}, especially in ROUGE-2 and METEOR. In contrast, the Pointer Generator Network with coverage mechanism \textbf{(PGNwithCOV)} consistently recorded the lowest scores across all evaluation metrics.

\begin{table*}[h!]
\centering
\begin{tabular}{|cccccc|} 
\hline
Model Name & ROUGE-1 & ROUGE-2 & ROUGE-L & METEOR &BERTScore\\ \hline
BART-base & \textbf{0.37} & \textbf{0.22} & \textbf{0.33} & 0.34 & \textbf{0.88} \\
T5-base & \textbf{0.37} & 0.21 & 0.32 & \textbf{0.39} & \textbf{0.88} \\
PEGASUS-X-base & 0.30 & 0.17 & 0.26 & 0.33 & 0.87 \\
ChatGPT-4 & 0.29 & 0.13 & 0.23 & 0.35 & 0.87 \\
LLaMA-3-8B (without fine-tuning) & 0.14 & 0.05 & 0.12 & 0.13 & 0.82 \\
LLaMA-3-8B (fine-tuned) & 0.27 & 0.11 & 0.25 & 0.23 & 0.87 \\
PGNwithCOV & 0.13 & 0.04 & 0.11 & 0.11 & 0.81 \\ 
\hline
\end{tabular}
%\caption{Comparison of Rouge-1, Rouge-2, Rouge-L, METEOR and BERTScore across BART-base, T5-base, PEGASUS-x-base, ChatGPT-4, LLaMA-3-8B (without and with fine-tuning) and PGNwithCOV models.}
\caption{Performance comparison of BART-base, T5-base, PEGASUS-x-base, ChatGPT-4, LLaMA-3-8B (with and without fine-tuning), and PGNwithCOV models across ROUGE-1, ROUGE-2, ROUGE-L, METEOR, and BERTScore metrics.}
\label{tab:results}
\end{table*}

\subsection{Comparison with Previous Works}

Table~\ref{tab:comparison} presents a comparison between our fine-tuned transformer-based models and competitive baselines, including LexRank \cite{xie2023survey} and the Posterior-and-Prior Knowledge Exploring-and-Distilling method (PPKED) \cite{liu2021exploring}, evaluated on the MIMIC-CXR dataset. As shown in Table~\ref{tab:results}, both fine-tuned \textbf{BART-base} and \textbf{T5-base} models achieved the highest ROUGE-1 score of 0.37 and a BERTScore of 0.88. The fine-tuned \textbf{BART-base} model further attained the best ROUGE-2 and ROUGE-L scores, with 0.22 and 0.33, respectively. Meanwhile, the fine-tuned \textbf{T5-base} model recorded the highest METEOR score of 0.39, reflecting strong semantic alignment with the reference impressions.

In contrast, \textbf{PPKED} underperformed in ROUGE-L (0.28 vs. 0.33) and recorded a substantially lower METEOR score of 0.15, suggesting limitations in the fluency and coherence of the generated impressions. \textbf{LexRank}, a traditional extractive summarization method, also lagged behind across all reported ROUGE metrics, achieving scores of 0.18 (ROUGE-1), 0.07 (ROUGE-2), and 0.16 (ROUGE-L). These findings underscore the advantages of employing modern, fine-tuned transformer-based models for generating accurate and coherent medical impressions.
\begin{table*}[h!]
\centering
\begin{tabular}{|cccccc|}
\hline
Model Name & ROUGE-1 & ROUGE-2 & ROUGE-L & METEOR & BERTScore \\
\hline
LexRank~\cite{xie2023survey} & 0.18 & 0.07 & 0.16 & -- & -- \\
PPKED~\cite{liu2021exploring} & -- & -- & 0.28 & 0.15 & -- \\
fine-tuned BART-base (ours) & \textbf{0.37} & \textbf{0.22} & \textbf{0.33} & 0.34 & \textbf{0.88} \\
fine-tuned T5-base (ours) & \textbf{0.37} & 0.21 & 0.32 & \textbf{0.39} & \textbf{0.88} \\ \hline
\end{tabular}
\caption{Performance comparison between our proposed models and existing baselines (LexRank and PPKED) on the MIMIC-CXR dataset using ROUGE, METEOR, and BERTScore metrics.}
\label{tab:comparison}
\end{table*}

\subsection{Case Study}
To access deeper insights into model behavior and the quality of the generated impressions, we present two case studies in Table~\ref{tab:case1} and Table~\ref{tab:case2}.
These examples illustrate how closely the \textit{model-generated impressions} align with the \textit{ground-truth impressions} from the MIMIC-CXR dataset.
For clarity, we use highlight colors: \hlr{pastel pink} highlights words or phrases generated by the model that are either irrelevant or factually incorrect.

Table~\ref{tab:case1} shows the first example, where fine-tuned LLaMA-3-8B and PEGASUS generate the most accurate \textit{impressions}, capturing key findings like ``pulmonary vascular congestion'' and ``atelectasis'' with appropriate uncertainty. ChatGPT-4 includes relevant details but also adds non-essential content such as ``tracheostomy tube in place with stable cardiomediastinal silhouette'' and ``PICC tip visibility'', which are not central to the \textit{impression}. BART and T5 produce concise outputs but miss critical information. The non-fine-tuned LLaMA-3-8B introduces a factual error, and PGNwithCOV fails with repetitive output like ``cardiopulmonary vascular...''.

\begin{table*}[ht!]
\centering
\begin{tabular}{|p{12.2cm}|}
\hline
\textbf{Input (\textit{findings}):} ``Single portable view of the chest is compared to previous exam from \_\_\_. Tracheostomy tube is again noted. Left PICC tip is not clearly delineated on the current exam. Again there is mild pulmonary vascular congestion. Streaky opacities at the lung bases suggestive of atelectasis; however infection cannot be excluded. Cardiomediastinal silhouette is stable as are the osseous and soft tissue structures.'' \\ \hline
\textbf{Ground-truth \textit{impression}:} ``No significant interval change since prior. Pulmonary vascular congestion. Bibasilar opacities potentially due to atelectasis; however, infection is not excluded.'' \\ \hline
\textbf{BART:} ``Streaky opacities at the lung bases suggestive of atelectasis.'' \\ \hline
\textbf{PEGASUS:} ``1. Mild pulmonary vascular congestion. 2. Streaky opacities at the lung bases suggestive of atelectasis; however infection cannot be excluded.'' \\ \hline
\textbf{T5:} ``Streaky opacities at the lung bases suggestive of atelectasis; however infection cannot be excluded.'' \\ \hline
\textbf{ChatGPT-4:} ``Tracheostomy tube in place with stable cardiomediastinal silhouette. Mild pulmonary vascular congestion and streaky opacities at lung bases suggestive of possible atelectasis or infection. Left PICC tip not clearly visualized.'' \\ \hline
\textbf{LLaMA-3-8B (without fine-tuning):} ``\hlr{No acute cardiopulmonary abnormality} is seen on this portable chest x-ray. Atelectasis is present.'' \\ \hline
\textbf{LLaMA-3-8B (fine-tuned):} ``Mild pulmonary vasocongestion with streaky bibasilar opacity, potentially atelectasis, but infection is difficult to exclude in the correct clinical setting.'' \\ \hline
\textbf{PGNwithCOV:} ``\hlr{No acute cardiopulmonary cardiopulmonary vascular vascular vascular vascular vascular vascular vascular vascular vascular vascular vascular}'' \\ \hline
\end{tabular}
\caption{Input findings and ground-truth impression from the test split of the MIMIC-CXR dataset. Outputs generated by seven models are presented for comparison.}
\label{tab:case1}
\end{table*}

Table~\ref{tab:case2} presents another example where T5 and ChatGPT-4 most closely match the ground-truth \textit{impression}, accurately identifying ``pulmonary edema'' improvement and ``bibasilar opacities''. ChatGPT-4 includes clinically correct but peripheral details such as ``tracheostomy'' and ``unchanged silhouette''. PEGASUS is generally faithful but verbose. BART captures pulmonary edema but omits bibasilar opacities, missing one of the key findings. The base LLaMA-3-8B hallucinates a contradiction by stating ``no acute cardiopulmonary abnormality'', while its fine-tuned version shows slight improvement. PGNwithCOV again fails due to decoding issues with repetitive <UNK> tokens and incorrect generation of ``No acute cardiopulmonary..''.

\begin{table*}[ht!]
\centering
\begin{tabular}{|p{12.2cm}|}
\hline
\textbf{Input (\textit{findings}):} ``Portable AP chest radiograph is obtained with the patient in the semi-erect position. Tracheostomy noted. Cardiomediastinal silhouette is unchanged; bulging of the pulmonary outflow tract reflects enlargement of pulmonary arteries and suggests underlying pulmonary arterial hypertension. Pulmonary edema has slightly improved compared to the prior study. Small right pleural effusion is unchanged. 
 Again bibasilar opacifications are noted and are suggestive of atelectasis or consolidation.'' \\ \hline
\textbf{Ground-truth \textit{impression}:} ``1. Unchanged bibasilar opacities are consistent with atelectasis or consolidation and pneumonia should be considered in the appropriate clinical context. 2. Improved pulmonary edema.'' \\ \hline
\textbf{BART:} ``Pulmonary edema has improved compared to the prior study. Small right pleural effusion is unchanged.'' \\ \hline
\textbf{PEGASUS:} ``Cardiomediastinal silhouette is unchanged; bulging of the pulmonary outflow tract reflects enlargement of pulmonary arteries and suggests underlying pulmonary arterial hypertension. Pulmonary edema has slightly improved compared to the prior study. Small right pleural effusion and bibasilar opacifications are suggestive of atelectasis or consolidation.'' \\ \hline
\textbf{T5:} ``1. Pulmonary edema has slightly improved compared to the prior study. 2. Bibasilar opacifications suggestive of atelectasis or consolidation.'' \\ \hline
\textbf {ChatGPT-4:} ``Tracheostomy in place with unchanged cardiomediastinal silhouette and signs of pulmonary arterial hypertension. Slight improvement in pulmonary edema from prior study, persistent small right pleural effusion, and bibasilar opacifications suggestive of atelectasis or consolidation.'' \\ \hline
\textbf{LLaMA-3-8B (without fine-tunned):} ``\hlr{No acute cardiopulmonary abnormality} is identified. Enlarged pulmonary vasculature is consistent with pulmonary hypertension, which may be secondary to chronic obstructive pulmonary disease.'' \\ \hline
\textbf{LLaMA-3-8B (fine-tuned):} ``Slight improvement in pulmonary congestion.'' \\ \hline
\textbf{PGNwithCOV:} ``\hlr{No acute cardiopulmonary cardiopulmonary <UNK> <UNK> <UNK> <UNK> <UNK> <UNK> <UNK> <UNK> <UNK> <UNK> <UNK>.}'' \\ \hline
\end{tabular}
\caption{Input findings and ground-truth impression from the test split of the MIMIC-CXR dataset. Outputs generated by seven models are presented for comparison.}
\label{tab:case2}
\end{table*}

\subsection{Human Evaluation}

To support the automated evaluation results with more detailed insights, a manual review was carried out. This human evaluation helped to capture aspects of summary quality—such as clarity, relevance, and clinical meaning—that automatic methods might miss.

Four human annotators, all pursuing a master's degree and having knowledge in natural language processing (NLP), took part in this process. Each annotator was given 10 randomly selected test examples. To avoid any bias, they were not informed about the names or internal details of the models. For each example, they were shown the original radiology \textit{findings} (input text), the correct \textit{impression} written by a radiologist (reference summary), and the \textit{impression} generated by different models (predicted summary): \textbf{T5-base}, \textbf{Pegasus-x-base}, \textbf{BART-base}, \textbf{ChatGPT-4}, \textbf{LLaMA-3-8B}, and \textbf{PGNwithCOV}.

The annotators were asked to place each model-generated summary into one of three categories: \textit{most preferred}, \textit{moderately preferred}, or \textit{least preferred}. They made their choices based on how accurate, useful, brief, and clear the summaries were.

\begin{table}[h!]
\centering
\scalebox{1} {
\begin{tabular}{|p{3.5cm}|p{3.5cm}|p{3.5cm}|}
\hline
Most Preferred & Moderately Preferred & Least Preferred\\ \hline
T5 & & \\
BART & T5 & PGNwithCOV\\
ChatGPT-4 & & \\
\hline
\end{tabular}
}
\caption{Top models in each preference category, according to human evaluation of the  summarization models across 10 sampled test cases.}
\label{tab:Human_Evaluation}
\end{table}

Table~\ref{tab:Human_Evaluation} shows that the results of the human evaluation highlight distinct preferences across the models tested. The \textbf{T5-base} was chosen as the \textit{most preferred} model in 3 out of 10 instances and 5 times as the \textit{moderately preferred} category. This suggests that T5 is effective in generating summaries that are clear, clinically meaningful, and structurally sound.

Both \textbf{BART-base} and \textbf{ChatGPT-4} also performed competitively, each earning the \textit{most preferred} label in 3 examples. \textbf{BART-base }was also chosen 2 times as the \textit{moderately preferred} category. It is noteworthy that \textbf{ChatGPT-4} was never placed in the \textit{least preferred} category, reflecting its consistent ability to generate fluent and accurate summaries regardless of the input variation.

The \textbf{Pegasus-x-base} model received a mixed response. It was marked as \textit{most preferred} only once and appeared 2 times in the \textit{moderately preferred} category. This indicates occasional success, though its performance lacked stability across samples.

In contrast, the \textbf{PGNwithCOV} achieved consistently low ratings. It was selected as the \textit{least preferred} in all 10 test cases, indicating a substantial gap in output quality compared to more advanced transformer-based architectures.

The \textbf{PGNwithCOV} was consistently selected as the \textit{least preferred model} by all evaluators across all 10 samples, indicating a clear preference for the more recent transformer-based models.

Overall, \textbf{T5-base}, \textbf{BART-base}, and \textbf{ChatGPT-4} each accounted for 30\% of the selections in the \textit{most preferred} category. \textbf{T5-base} emerged as the leading model in the \textit{moderately preferred} category, being chosen in 50\% of the cases. The \textbf{PGNwithCOV} was consistently placed in the \textit{least preferred} category by all evaluators across all 10 samples.

\section{Conclusion and Future Scope}

In this study, we showed that fine-tuned transformer models—particularly BART-base, T5-base, and Pegasus-x-base—are effective in summarizing radiology \textit{findings} into \textit{impressions}. Our findings also underscore the promise of large language models (LLMs) in generating high-quality summaries, supporting their potential use in real-world clinical settings. Human evaluation complemented automated metrics by capturing important qualitative aspects such as readability, coherence, and clinical relevance.

Although our evaluation focused on standard metrics such as ROUGE, METEOR, and BERTScore, we acknowledge the need for clinical validation to assess the practical utility of the generated impressions. In future work, we aim to collaborate with medical professionals to obtain expert feedback on the accuracy, clarity, and clinical applicability of the outputs. This step will be crucial for ensuring model reliability and enabling their integration into clinical workflows.

\bibliographystyle{unsrt}
\bibliography{ref, anthology}
\end{document}